%% file: main.tex
\documentclass[conference]{IEEEtran}
\IEEEoverridecommandlockouts
\usepackage{cite}
\usepackage{amsmath,amssymb,amsfonts}
\usepackage{algorithmic}
\usepackage{graphicx}
\usepackage{textcomp}
\usepackage{xcolor}
\usepackage{algorithm}
\usepackage[square,sort,numbers]{natbib}

\makeatletter
\def\ps@IEEEtitlepagestyle{%
  \def\@evenfoot{}%
}

\def\BibTeX{{\rm B\kern-.05em{\sc i\kern-.025em b}\kern-.08em
    T\kern-.1667em\lower.7ex\hbox{E}\kern-.125emX}}

\begin{document}

\title{GCN-SE: Attention as Explainability \\ for Node Classification in Dynamic Graphs
}

 \author{\IEEEauthorblockN{Yucai Fan}
 \IEEEauthorblockA{
 \textit{Carnegie Mellon University}\\
 fanyucaiharold@gmail.com}
 \and 
 \IEEEauthorblockN{Yuhang Yao}
 \IEEEauthorblockA{
 \textit{Carnegie Mellon University}\\
 yuhangya@andrew.cmu.edu}
 \and 
 \IEEEauthorblockN{Carlee Joe-Wong}
 \IEEEauthorblockA{
 \textit{Carnegie Mellon University}\\
 cjoewong@andrew.cmu.edu}
 }

\maketitle

\begin{abstract}
Graph Convolutional Networks (GCNs) are a popular method from graph representation learning that have proved effective for tasks like node classification tasks. Although typical GCN models focus on classifying nodes within a static graph, several recent variants propose node classification in dynamic graphs whose topologies and node attributes change over time, e.g., social networks with dynamic relationships, or literature citation networks with changing co-authorships. These works, however, do not fully address the challenge of flexibly assigning different importance to snapshots of the graph at different times, which depending on the graph dynamics may have more or less predictive power on the labels. We address this challenge by proposing a new method, GCN-SE, that attaches a set of learnable attention weights to graph snapshots at different times, inspired by Squeeze and Excitation Net (SE-Net). We show that GCN-SE outperforms previously proposed node classification methods on a variety of graph datasets. To verify the effectiveness of the attention weight in determining the importance of different graph snapshots, we adapt perturbation-based methods from the field of explainable machine learning to graphical settings and evaluate the correlation between the attention weights learned by GCN-SE and the importance of different snapshots over time. These experiments demonstrate that GCN-SE can in fact identify different snapshots' predictive power for dynamic node classification. 

\end{abstract}


\input{introduction}

\input{related}

\input{method}


\input{experiments}

\section{Conclusion}\label{sec:conclusion}
In this paper, we first proposed a new method, GCN-SE, to predict nodel labels in a dynamic graph based on aggregation of snapshots and node attributes at different timesteps. Inspired by SE-net, which proposes Squeeze and Excitation blocks, we use learned attention weights corresponding to different timesteps in aggregating the different snapshots. We also make use of attribute dynamics by applying separate attention weights to these attributes. Experiments on real-world data sets show that GCN-SE outperforms most baseline models on the metrics of accuracy, AUC and F1 score. The attention weights also provide a way to give interpretative capacity to the model, by signaling the importance of snapshots at different time steps. Inspired by perturbation-based methods in interpretable machine learning, we rigorously define the importance of a timestep as the change in accuracy after masking the attention in that timestep. We propose a correlation metric to measure the relevance between our definition of importance and our attention weights, and use this metric to verify the effectiveness of GCN-SE's attention weights as an explainatory tool. We finally show that these weights can be used to detect various temporal patterns that may occur in real-world datasets.

\section*{Acknowledgements}
This work was partially supported by NSF CNS-1909306.
\bibliographystyle{IEEEtran}
\bibliography{ref}

\end{document}

%% file: introduction.tex
\section{Introduction}
Over the past several years, (deep) neural networks have demonstrated considerable success at tackling machine learning problems ranging from image recognition to time series prediction, among other areas. Recent efforts have extended deep learning's success to graphically structured data, such as social networks where each node represents a user and edges represent relationships between them. Graph neural networks (GNNs) can then be used to solve problems such as node classification~\cite{kazemi2020representation}, which assumes that each node in a graph is associated with a label, e.g., in our social network example, the labels might be the city in which the user lives, or the political party to which a user belongs. The node classification problem is then to predict the label of another user in the social network, given knowledge of the edges connecting the user to others and potentially other features (like the age of the user) attached to each node in the graph.
We can frame the node classification problem as a graph-based semi-supervised learning problem, where at training time we only know the labels of a subset of the graph nodes. The challenge is then to design and train a GNN that takes as input abstracted information about the graph structure around each node, and outputs predictions of each node's class label.

Graph Convolutional Networks (GCN) are one of the most widely used methods to abstract graph information for node classification tasks. A GCN consists of multiple graph convolutional layers, each of which transforms the current node representation into another, summarizing the graph structure around a node so as to make it useful for predicting the node label. Many prior works focus on node classification in \emph{static} graphs, where the node topology and labels do not change. However, in practice both the node labels and graph topology may change~\cite{yao2020interpretable,xu2019adaptive}: in our social network example above, users may move to different places (changing their labels) and form new friendships (changing the graph topology), potentially requiring us to retrain the GCN as the pattern of relationships between the labels and topology changes. Indeed, dynamic graphs are found in many applications of node classification: in most social networks, for example, users will form new connections over time and may migrate between groups. 

Node classification on a dynamic graph focuses on predicting the node labels at a specific timestep, given the history of the graph structure at previous timesteps. Prior work has identified three main \textbf{challenges} of node classification for dynamic compared to static graphs~\cite{xu2019adaptive}: (i) \emph{entangled} spatial and temporal information, (ii) the \emph{evolution} of both node attributes and the graph topology, and (iii) the \emph{variation} in the effect of factors like different snapshots or models on node representations for different datasets. Since the importance of these various factors is generally not apparent a priori for specific datasets, it is then desirable for a node classification method to also give information on their relative importance to the predictions. For example, classification methods for graphs whose topology changes rapidly may emphasize recently formed edges, while graphs that exhibit temporal periodicity would need classification methods that emphasize certain past timesteps~\cite{yao2020interpretable}. Proper classification methods should discover and exploit such temporal patterns in predicting node labels.

Several prior works on node classification have attempted to address these challenges by crafting a summary of the graph topology across all prior timesteps as the input of the prediction model, or by explicitly modeling the temporal evolution of the node attributes and graph topology~\cite{hisano2018semi}. These methods can handle the first challenge of entangled spatial and temporal information and the second challenge of the evolving node attributes and graph topology. However, they must be flexible enough to fully capture the variation in graph evolution patterns across different datasets and thus different types of graphs~\cite{xu2019adaptive}. 
Some prior work has proposed deep learning methods to learn flexible models~\cite{sankar2018dynamic,pareja2020evolvegcn}, but many of these black-box methods do not provide much insight into which features or temporal patterns they are exploiting in making their predictions. Even those that attempt to provide such insights (generally with some form of attention~\cite{xu2019adaptive}) rarely verify that their metrics for feature importance are in fact correlated with the model's explanatory power, and do not evaluate the significance of the graph topology at different times~\cite{xu2019adaptive}. Prior work in interpretable machine learning has shown that attention may not be a good indicator of feature importance~\cite{serrano2019attention}, which calls into question the true interpretability of these models. We address these challenges in this work.


In this paper, we propose GCN-SE (GCN-Squeeze and Excitation), a new graph node classification method that employs the widely known concept of \emph{attention} from deep learning to improve the performance of existing methods while explaining the importance of information collected at different timesteps. In doing so, we address the third challenge presented above: by learning the attention weights, we ensure that our model is flexible enough to perform well on multiple types of graphs. We also address a \textbf{new challenge}: existing methods to evaluate feature importance scores often rely on the ability to perturb features like specific pixels in image inputs~\cite{bhatt2020explainable,lundberg2017unified}, which cannot be directly applied to our dynamic graph inputs. We devise appropriate perturbations for graphical data and measure their effects on the model predictions, enabling us to precisely quantify the explanatory power of our attention-based architecture in signaling the importance of different factors (in particular, the graph structure at different times) to the node classification results.
%
We summarize our \textbf{three major contributions} as follows:
\begin{itemize}
    \item We propose \textbf{\textit{GCN-SE}}, which uses attention methods to solve dynamic node classification problems. The attention weights are learned as the weights of a linear combination of node representations over time, and attribute dynamics can be also taken into consideration;
    \item We propose \textbf{\textit{novel metrics}} to evaluate the correlation between the importance of different timesteps and the importance indicated by the learned attention weights, thus quantifying the explanatory power of a given classification model;
    \item We finally \textbf{\textit{verify}} that (i) GCN-SE outperforms existing node classification methods on dynamic graph datasets, and (ii) our use of attention can serve as an explanatory tool to the model and data, according to the metrics defined in our second contribution.
\end{itemize}

We contrast our approach with related work in Section~\ref{sec:related} before introducing our proposed GCN-SE method in Section~\ref{sec:method}. We then evaluate the performance of GCN-SE on several network datasets in Section~\ref{sec:experiment}, and show that our learned attention weights can in fact explain GCN-SE's prediction results in Section~\ref{sec:explanation}. We conclude and present some directions for future work in Section~\ref{sec:conclusion}.

%% file: related.tex
\section{Related Work}\label{sec:related}

\textbf{Feature aggregation methods.} A natural method to handle dynamic graphs is to aggregate snapshots of the graph at different timesteps, which we may also call subgraphs. This aggregated snapshot can then be viewed as a static graph representation that is fed into a classification model, just as for static graphs.
An early example~\cite{liben2007link} proposed a simple aggregation method that added up the graph adjacency matrices at each timestep and applied a static decoder to the sum to perform predictions. Similarly, ~\cite{sharan2008temporal} takes an average of the adjacency matrices at each timestep. In~\cite{hisano2018semi}, the authors define a similar aggregation, but more explicitly capture the temporal evolution of the graph by defining a formation matrix and dissolution matrix considering the edges having been added or removed since the previous snapshot. 
More recent proposals give more weight to more recent snapshots when node features evolve over time~\cite{ahmed2016efficient}, for example by first applying a static feature extractor and then aggregating the results over time, or aggregating features by taking a weighted average with exponentially decaying weights~\cite{zhu2012hybrid,yao2016link}. 
One can also learn different exponential decay weights for different nodes in the graph depending on their labels~\cite{yao2020interpretable}, which may also provide tools to add interpretability to the model.

\textbf{Recurrent model architectures.} Aggregation methods permit an easy translation of prediction methods for static graphs to dynamic graphs, allowing us to apply the many known methods for static graphs to dynamic ones. However, they raise a natural question of \emph{how best to aggregate snapshots of the graph at different timesteps}. When using exponentially decaying weighted averages, for example, choosing the correct weight can significantly affect the prediction result~\cite{yao2020interpretable}.

Many works attempt to learn the optimal aggregation method by using a recurrent GNN architecture: a GNN, such as a graph convolutional network (GCN), can act as a feature extractor, while the temporal dynamics of the extracted features are learned with the recurrent architectures such as RNN (recurrent neural network) or LSTM (long-short term memory) structures.
For example,~\cite{velivckovic2017graph} proposed graph attention networks (GAT), leveraging masked self-attention layers to highlight different node features. In~\cite{seo2018structured}, the authors propose to learn the dynamics by either using GCN to obtain node embeddings that are fed into a LSTM, or taking node features as input but replacing fully connected layers in LSTM with graph convolutions. EvolveGCN~\cite{pareja2020evolvegcn} uses a RNN to evolve the GCN parameters, thus capturing the evolution of the graph topology. These methods, however, are generally not interpretable due to their use of GCNs to learn new representations of the graph topology and node features. 

\textbf{Attention methods and interpretability.} Attention methods in particular have become a popular component of dynamic graph representation learning methods in recent years. DynSAT ~\cite{sankar2018dynamic} proposes to compute node representations by employing self-attention to capture both structural properties and temporal patterns, while ~\cite{zheng2019addgraph} combines GCN and a contextual attention-based model to perform anomaly detection. ~\cite{xu2019spatio} uses attention to provide interpretable weights to node features, but not to snapshots of the graph topology taken at different times. 
Many of these methods claim that the use of attention allows them to better interpret the significance of each feature to the model prediction. None of them, however, verify this assertion, though prior work has called into question the use of attention to signify feature importance~\cite{serrano2019attention}. In this work, we propose a novel attention-based method, GCN-SE, that applies attention to snapshots of the node topology, as well as node features, at different times. We further verify that GCN-SE's attention weights can be interpreted as the significance of different timesteps. 




%% file: method.tex
\section{Introducing GCN-SE}\label{sec:method}
This section presents Graph Convolutional Network with Squeezing and Excitement (GCN-SE) for node classification in dynamic graphs. We first formulate the problem, propose our model and method, and then discuss training details.

\subsection{Problem Definition}

\begin{figure}[ht]
 	\centering
 	\includegraphics[width=0.4\textwidth]{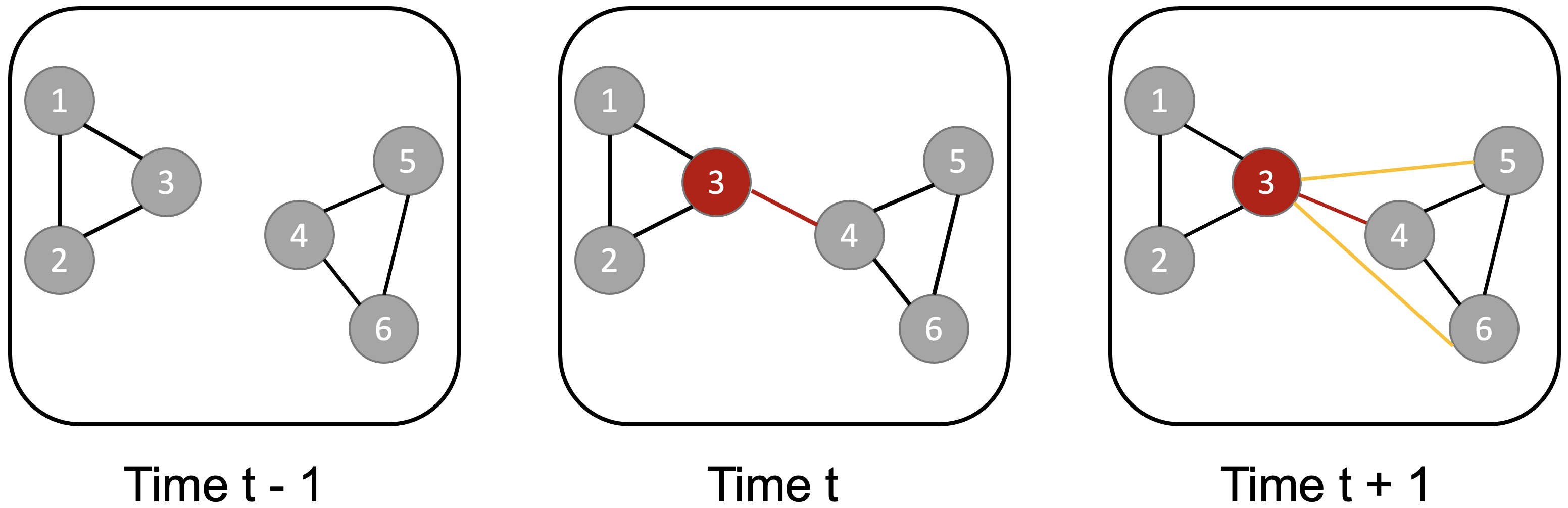}
 	\caption{Dynamic Graph with Changing Classes. At timestep $t-1$, nodes $1,2,3$ and nodes $4,5,6$ respectively belong to two different classes. Node $3$ changes the class membership at timestep t and forms new edges with the members in the new class (node $4$ at timestep $t$; node $5$ and $6$ at timestep $t+1$).}
 	\label{fig_change_membership} 
 \end{figure}

Suppose $T$ to be the total number of timesteps; we use $t = 1,2,\ldots,T$ to denote each timestep. A dynamic graph can be defined as an ordered sequence of snapshots or subgraphs: $S = {G_1, G_2, ..., G_T}$ with $G_t = (V, \mathbf{A}_t, \mathbf{X}_t, \mathbf{\Theta}_t)$ at each timestep $t$ denoting the snapshot of timestep $t$. Here $V$ is the set of nodes (which we assume to be constant over time, as in the social network and citation datasets we consider in Section~\ref{sec:experiment}), $\mathbf{A}_t$ denotes the adjacency matrix between these nodes, only including edges formed at timestep $t$, $\mathbf{X}_t$ is a feature matrix of the attributes associated with each node at time $t$, and $\mathbf{\Theta}_t$ denotes the class membership matrix (i.e., class labels for each node in $V$), which can change over time. Some graphs may not have attribute matrices, in which case $\mathbf{X}_t$ may be omitted. At timestep $T$, given the dynamic graph $S$ with the incomplete membership matrix $\mathbf{\Theta}_T^{train}$ that associates a  subset of ``training'' nodes with known class labels at time $T$, our task is to classify the remaining nodes with unknown class labels, i.e., to estimate the membership matrix $\mathbf{\hat{\Theta}}_T$. Figure~\ref{fig_change_membership} illustrates an example dynamic graph with six nodes, where one node changes its class and forms new edges at time $t$.

\subsection{Graph Convolutional Network}

Graph convolutions are generalized convolution for graphs, and are widely used for graph representation learning tasks such as edge prediction, node classification, etc. Denoting their parameter matrix as $\mathbf{W}$, the $l$-th graph convolutional layer in a graph convolutional network (GCN) takes node representation $\mathbf{Z}_{l}$ as input and outputs a transformed representation $\mathbf{Z}_{l+1}$. Many designs of transform functions have been proposed (see \cite{zhou2020graph}), and here we make use of a typical design, which is proposed by \cite{kipf2016semi} and formulated as:

\begin{equation}
    \mathbf{Z}_{l+1}=\sigma\left(\tilde{\mathbf{D}}^{-\frac{1}{2}} \tilde{\mathbf{A}} \tilde{\mathbf{D}}^{-\frac{1}{2}} \mathbf{Z}_{l} \mathbf{W}_{l+1}\right)
\end{equation}

Here $\sigma$ is a non-linear function (typically ReLU, with a softmax in the last layer), $\tilde{\mathbf{A}}=\mathbf{A}+\mathbf{I}_{N}$, $\mathbf{A}$ is the adjacency matrix, $\mathbf{I}_{N}$ is the identity matrix, and $\mathbf{D}$ is the degree matrix, defined as $\widetilde{\mathbf{D}}_{ii}=\sum_{j} \tilde{\mathbf{A}}_{i j}$. For graphs with attribute matrices $\mathbf{X}$, $\mathbf{Z}_{0}$ can be initialized to $\mathbf{X}$ where the rows in $\mathbf{X}$ stand for the attribute values of the nodes. For graphs without node attributes, $\mathbf{Z}_{0}$ can be initialized to an identity matrix where the rows are a one-hot encoding of the nodes. 

\subsection{Squeeze and Excitation}
We extend the GCN idea to handle dynamic graphs by considering the snapshot at each timestep to be a ``channel'' of data, analogous to the RGB color channels in image datasets. We then use Squeeze-and-Excitation (SE) to capture the temporal information; we introduce the SE idea here and explain its use in our proposed GCN-SE method in the next subsection. Squeeze-and-Excitation is first proposed in \cite{hu2018squeeze} to tackle the presence of multiple channels in training datasets. The SE block consists of two parts: squeeze and excitation. The ``squeeze'' operation aims to exploit channel dependencies. In order to squeeze global spatial information into a channel descriptor, the SE block uses global average pooling to generate channel-wise statistics, represented by a vector 
$\mathbf{z} \in \mathbb{R}^{C}$. Here $\mathbf{u}_c$ is the feature map input, $C$ is the number of channels, $H$ and $W$ define the height and width of the feature maps, and each element $c$ of $\mathbf{z}$ is generated by:

\begin{equation}
    \mathbf{z}_{c}=\mathbf{F}_{s q}\left(\mathbf{u}_{c}\right)=\frac{1}{H \times W} \sum_{i=1}^{H} \sum_{j=1}^{W} \mathbf{u}_{c}(i, j).
\end{equation}

The purpose of ``excitation'' operations is to make use of the aggregated information produced by the squeeze operations, which learn channel-wise dynamics, and can be considered as the extent to enhance the channel information. Suppose $\delta$ refers to the ReLU function, $\mathbf{W_1} \in \mathbb{R}^{L \times C}$ and $\mathbf{W_2} \in \mathbb{R}^{C \times L}$, where $L$ is the size of the hidden layer output. The excitation is then a gating mechanism with a sigmoid activation $\phi$: 

\begin{equation}
   \mathbf{s}=\mathbf{F}_{e x}(\mathbf{z}, \mathbf{W})=\phi(g(\mathbf{z}, \mathbf{W}))=\phi\left(\mathbf{W}_{2} \delta\left(\mathbf{W}_{1} \mathbf{z}\right)\right), 
\end{equation}
and the final output of this squeeze-and-excitation block is 

\begin{equation}
    \overline{\mathbf{x}}_{c}=\mathbf{F}_{\text {scale }}\left(\mathbf{u}_{c}, s_{c}\right)=s_{c} \mathbf{u}_{c}
\end{equation}
where the output $\overline{\mathbf{X}}=\left[\overline{\mathbf{x}}_{1}, \overline{\mathbf{x}}_{2}, \ldots, \overline{\mathbf{x}}_{C}\right]$ is the enhanced feature map from the original input.

\subsection{GCN-SE architechture}
\begin{figure}[ht]
	\centering
	\includegraphics[width=0.45\textwidth]{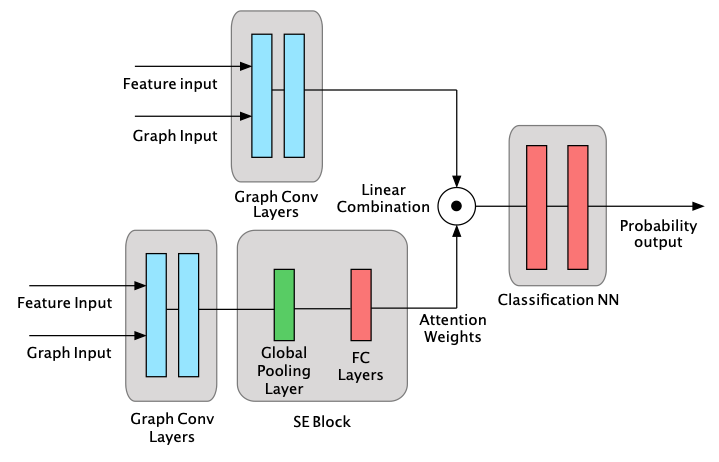}
	\caption{GCN-SE model structure. The outputs of the graph convolution layers for each timestep are fed into the SE block to generate attention weights. These weights are then used to take a linear combination of the convolution layer outputs, which are then used to make the classification prediction.}
	\label{fig_method}
\end{figure}

We now introduce the architecture of our proposed method, GCN-SE. As shown in Figure~\ref{fig_method}, GCN-SE consists of two blocks of graph convolutional layers, which act as feature extractors. The inputs of these layers are the adjacency matrix and node attributes as node representations over time, which are transformed into feature maps by the graph convolutions. Unlike prior work that approaches node classification from a traditional recurrent network perspective, we do not explicitly model the temporal graph evolution and instead regard the adjacency matrices $\mathbf{A}_1, \mathbf{A}_2, ...,\mathbf{A}_T$ at different timesteps as separate channels of data. The features extracted from each channel by the convolutional layers are then taken as inputs to a squeeze and excitation block, which is used to capture temporal dynamics and as explained above can be used for feature map aggregation.


Inspired by SE-net, we propose to use the SE block to learn a set of attention weights. As shown in Figure~\ref{fig_method}, these weights are used to calculate a linear combination of the features extracted from the convolutional layers, where the weight of the features for each timestep in the linear combination is given by the output of the SE block. 
Finally, the linear combination across timesteps is fed into a fully connected network to produce the final classification result.

The GCN-SE model training process is shown in Algorithm \ref{alg_GCNSE}. After the initialization (lines 1 to 3), we pass the inputs through two graph convolutional layers (lines 4 to 7). These are used to learn attention weights in squeeze and excitation operations in line 8 and line 11, respectively. In lines 12 to 15, we again pass the original inputs through a set of graph convolutional layers; we then take a linear combination of the result using the attention weights from the SE block (line 16) and pass the output through a fully connected network for classification (line 17). The training process uses the cross-entropy error (line 18) to backpropagate the weights (line 19), until obtaining the final classification result (line 21). 

\renewcommand{\algorithmicrequire}{ \textbf{Input:}} 
\renewcommand{\algorithmicensure}{ \textbf{Output:}} 
\begin{algorithm}
    \caption{GCN-SE}
    \label{alg_GCNSE}
    \begin{algorithmic}[1]
        \REQUIRE ~~ \\
            Adjacency matrices: $\mathbf{A}_1, \mathbf{A}_2, ..., \mathbf{A}_T$; \\
            Node attributes matrix: $\mathbf{X}_1, \mathbf{X}_2, ..., \mathbf{X}_T$; \\
            Number of nodes: $N$; \\
            Membership matrix for training data: $\mathbf{\Theta}^{train}_{T}$ 
        \ENSURE: ~~ \\
            Predicted probabilities for timestep $T$: $\mathbf{\hat{\Theta}}_{T}$ \\
            Attention weights: $\mathbf{W}_{att}$
        \FORALL{$t$ from 1 to T}
            \STATE Initialize $\mathbf{Z}_t^{(0)}=\mathbf{X}_t$ (or $\mathbf{I}_N$ if no attributes)
            available
        \ENDFOR    
        
        \FOR{iteration $i$ from 1 to MAX\_ITER}
            \FOR{$t$ from 1 to T}
                \STATE $\mathbf{Z}_t^{(1)} = \sigma\left(\tilde{\mathbf{D}}^{-\frac{1}{2}} \tilde{\mathbf{A}} \tilde{\mathbf{D}}^{-\frac{1}{2}} \mathbf{Z}_t^{(0)} \mathbf{W}^{(1)}\right)$
                \STATE $\mathbf{Z}_t^{(2)} = \sigma\left(\tilde{\mathbf{D}}^{-\frac{1}{2}} \tilde{\mathbf{A}} \tilde{\mathbf{D}}^{-\frac{1}{2}} \mathbf{Z}_t^{(1)} \mathbf{W}^{(2)}\right)$
                \STATE $c_t = Pooling(\mathbf{Z}_t^{(2)})$
            \ENDFOR
            
            \STATE $\mathbf{c} = \left(c_1, c_2, ..., c_T\right)$
            \STATE $\mathbf{W}^{att} = \phi\left(\mathbf{W}_{2} \delta\left(\mathbf{W}_{1} \mathbf{c}\right)\right)$
                    
            \FOR{$t$ from 1 to T}
                \STATE $\mathbf{\hat{Z}}_t^{(1)} = \sigma\left(\tilde{\mathbf{D}}^{-\frac{1}{2}} \tilde{\mathbf{A}} \tilde{\mathbf{D}}^{-\frac{1}{2}} \mathbf{Z}_t^{(0)} \mathbf{W}^{(1)}\right)$
                \STATE $\mathbf{\hat{Z}}_t^{(2)} = \sigma\left(\tilde{\mathbf{D}}^{-\frac{1}{2}} \tilde{\mathbf{A}} \tilde{\mathbf{D}}^{-\frac{1}{2}} \mathbf{\hat{Z}}_t^{(1)} \mathbf{W}^{(2)}\right)$
            \ENDFOR
    
            \STATE $\hat{\mathbf{Z}} = \Sigma^{T}_{t = 1} \left( \mathbf{W}_t^{att} \mathbf{\hat{Z}}_t^{(2)} \right)$
            
            \STATE $\mathbf{\hat{\Theta}}_T = \sigma \left( \text{fc}(\hat{\mathbf{Z}}) \right)$
            \STATE CrossEntropyLoss($\mathbf{\hat{\Theta}}^{train}_T$, $\mathbf{\Theta}^{train}_T$)
            \STATE Backward()
        \ENDFOR
        \STATE ${\mathbf{\Theta}}_{T}=\text { Onehot }\left(\arg \max _{1 \leq j \leq n} \mathbf{\hat{\Theta}}_T\right)$
    \end{algorithmic}
\end{algorithm}


We finally highlight an alternative to Algorithm~\ref{alg_GCNSE} that explicitly handles dynamic node attributes and assigns them attention weights that may correspond to their importance. 
To do so, we can separate the aggregation into two parts: aggregation for the adjacency matrix and aggregation for the node attributes, with two set of attention weights learned separately according to the process illustrated in Algorithm~\ref{alg_GCNSE}. 
We can thus disentangle the learning of the topology and attribute dynamics in predicting the node labels, which improves GCN-SE's performance as demonstrated in Section~\ref{sec:experiment}'s experimental results on real datasets.  

%% file: experiments.tex
\section{Experimental Results}\label{sec:experiment}

We next evaluate our proposed GCN-SE model by comparing its results with prevailing baseline methods on several real dynamic graph datasets. 

\subsection{Datasets}

We conducted experiments on five real datasets, as shown in Table \ref{tab:datasets}. All datasets have edges that form at different times although only nodes in DBLP-E change their class over time. The DBLP-3, DBLP-5, Reddit and Brain datasets also have dynamic node attributes. 

\begin{table}[h]
    \centering
    
    \begin{tabular}{|cccccc|}
    \hline
         Dataset& Nodes &Edges & Time Steps &Classes&Attributes \\
    \hline\hline
          DBLP-E & 6942 & 327392 & 14 &2 &$\times$\\
          DBLP-3 & 4257 & 23540 & 10 &3 &100 \\
          DBLP-5 & 6606 & 42815 & 10 & 5 &100\\
          Brain & 5000 & 1955488 & 12 &10 &20\\
          Reddit & 8291 & 264050 & 10 &4 &20\\
    \hline
    \end{tabular}
\vspace{1mm}
    \caption{Real datasets for evaluation of our methods.}
    \label{tab:datasets}
\end{table}

\textbf{DBLP-E \&  DBLP-3 \& DBLP-5} These datasets are extracted from DBLP\footnote{https://dblp.org/}, which provides a large collection of bibliographic information from major conferences and journals in various subfields of computer science. In these datasets, every author is extracted as a node in a graph, and co-authorship determines whether there is a connection between two nodes. DBLP-E~\cite{yao2020interpretable} considers 14 year-long timesteps, corresponding to the years 2005 to 2018, and creates a dynamic graph by defining the snapshot in each year based on the authorship of each paper published in that year. The node label in each year is defined as the primary research sub-field of the author, as indicated by the index term that appears most frequently in that author's papers in that year. The label may change from year to year as authors move between subfields. DBLP-E is constructed without attributes and only consists of graph topologies, while DBLP-3 \& DBLP-5 include node attributes extracted by \texttt{word2vec}~\cite{mikolov2013efficient} from the authors’ paper titles and abstracts in each year. The authors in DBLP-3 and DBLP-5 are clustered into three and five classes (corresponding to different research areas) respectively. These classes are static and do not change over time.

\textbf{Reddit} This dataset is generated from Reddit\footnote{https://www.reddit.com/}, which is known as a social news aggregation, web content rating, and discussion website. The graph is constructed by considering nodes as posts; two nodes are connected if they share keywords. The node attributes are also generated by \texttt{word2vec} on the post comments~\cite{hamilton2017inductive}.

\textbf{Brain} This dataset is generated from functional magnetic resonance imaging (fMRI) data\footnote{https://tinyurl.com/y4hhw8ro}. Nodes represent cubes of brain tissue, and two nodes are connected during a time period if they show similar degrees of activation during that time period. Node attributes are generated by principal component analysis (PCA) on the fMRI data.

\subsection{Baselines and Metrics}
We compare our model with multiple baselines. GCN~\cite{kipf2016semi} and GraphSage \cite{hamilton2017inductive} are supervised methods that include static graph structure and node attributes; they both ignore temporal information. GC-LSTM~\cite{chen2018gc} is another supervised method that utilizes the temporal information of both graphs and node attributes. RNNGCN~\cite{yao2020interpretable} employed a 2-layer GCN and introduced a decay weight as a learnable parameter; information from each timestep was multiplied by this weight, which decayed over time, and the resulting linear combination over time was used for classification. Thus, this method is similar to ours, but assumes that the importance of different timesteps decays exponentially for more distant timesteps, while we learn more flexible attention weights for different timesteps. We evaluate the performance of all methods with the standard accuracy (ACC), area under the ROC curve (AUC) and F1-score classification metrics.

\subsection{Experimental Settings}


We randomly divide each dataset into 70\% training / 20\% validation / 10\% test points. Each method uses two hidden Graph Neural Network layers (GCN, GraphSage, etc.) with the layer size equal to the number of classes in the dataset. We add a dropout layer between the two layers with dropout rate 0.5. We use the Adam optimizer with learning rate 0.0025. Each method is trained with 500 iterations. For our GCN-SE model, the hidden units of SE-net are set to half the number of channels ($r=0.5$). 

For static methods (GCN, GraphSage) we first accumulate the adjacency matrices of graphs at each timestep, and then use the normalized accumulated graph and the node attributes at the last timestep as inputs. RNNGCN takes the temporal graphs and attributes at the last timestep as inputs. For GC-LSTM and our proposed GCN-SE method, we use the temporal graphs and temporal node attributes as inputs. 
%
We implemented the models based on the Pytorch framework. 
The code of all methods and datasets are publicly available\footnote{https://github.com/GCN-SE/GCN-SE}.
\subsection{Results}
\begin{figure}[ht]%
    \centering
    {\includegraphics[width=7cm]{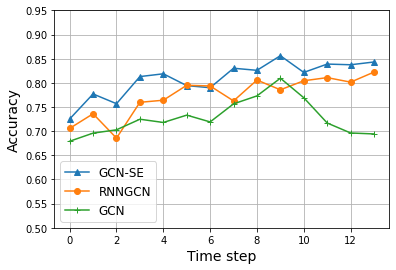} }%
    {\includegraphics[width=7cm]{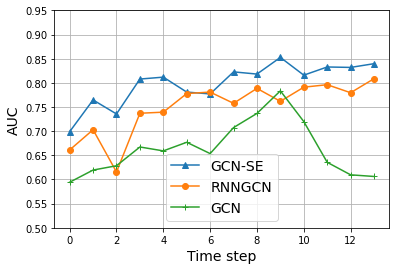} }%
    {\includegraphics[width=7cm]{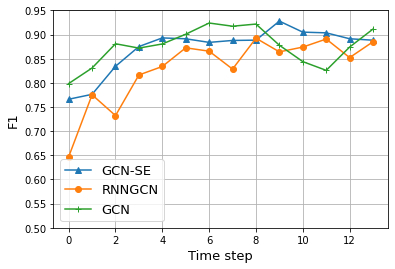} }%
    \caption{GCN-SE outperforms RNNGCN and GCN on the DBLP-E dataset.}%
    \label{fig_DBLPE}%
\end{figure}

Figure~\ref{fig_DBLPE} shows the accuracy, AUC and F1 score of the proposed method and baseline methods on each timestep of the DBLP-E dataset. Since node labels in DBLP-E may change over time, we predict the labels of each time step based on all previous graph snapshots. To demonstrate the effect of using different weights for snapshots across time, we take RNNGCN~\cite{yao2020interpretable} (which uses exponentially decaying weights) and GCN (which uses uniform weights) as baselines. The results indicate that GCN-SE achieved better results for accuracy and AUC across all timesteps, while the F1 scores are close for all three models. Since GCN is a static method, it generally achieves flatter accuracy and AUC over time, while GCN-SE and RNNGCN are both able to increase their performance over time as more prior snapshots become available.

We next consider the remaining datasets, whose labels are consistent overtime. We take RNNGCN, GraphSage, GCN, and GC-LSTM as our baselines. To evaluate the use of attribute dynamics in GCN-SE, we define the variants GCN-SE-1, whose attention weights are only calculated by snapshot topology and ignore attribute dynamics (Algorithm~\ref{alg_GCNSE}); and GCN-SE, which has two sets of attention weights calculated separately from the dynamic toplogy and dynamic attributes (see the discussion at the end of Section~\ref{sec:method}). Tables~\ref{tab_exp_ACC} to~\ref{tab_exp_F1} respectively show the accuracy, AUC and F1 scores of proposed method and baseline methods on the DBLP-3, DBLP-5, Reddit, and Brain datasets. On most datasets, our proposed GCN-SE outperforms the other baselines on all three metrics. All of the results are averaged over 30 runs.


\begin{table}[h!]
\centering
\begin{tabular}{|c c c c c|} 
 \hline
 & DBLP-3 & DBLP-5 & Reddit & Brain \\ [0.5ex] 
 \hline\hline
 GCN-SE & \textbf{0.7820} & \textbf{0.6762} & 0.3009 & \textbf{0.5096} \\
 GCN-SE-1  & 0.7674 & 0.6683 & 0.3044 & 0.4733\\
 RNNGCN & 0.7650  & 0.6692 & 0.2891 & 0.4148\\
 GraphSage& 0.7610 & 0.6614 & 0.3103 & 0.5090\\
 GCN & 0.7589 & 0.6678 & 0.2965 & 0.3818 \\
 GC-LSTM& 0.7674 & 0.6531 & \textbf{0.3434} & 0.4768\\ [1ex] 
 \hline
\end{tabular}
\vspace{1mm}
\caption{Accuracy (best method for each dataset bolded).}
\label{tab_exp_ACC}
\end{table}

\begin{table}[h!]
\centering
\begin{tabular}{|c c c c c|} 
 \hline
 & DBLP-3 & DBLP-5 & Reddit & Brain  \\ [0.5ex] 
 \hline\hline
 GCN-SE & \textbf{0.6912} & \textbf{0.5782} & 0.2151 & 0.4838  \\ 
 GCN-SE-1 & 0.6678 & 0.5570 & 0.1968 & 0.4541  \\ 
 RNNGCN & 0.6646  & 0.5498 & 0.1755 & 0.4026 \\
 GraphSage& 0.6618 & 0.5411 & 0.1580 & \textbf{0.4929} \\
 GCN & 0.6598 & 0.5411 & 0.1701 & 0.3332 \\
 GC-LSTM& 0.6813 & 0.5295 & \textbf{0.2789} & 0.4284 \\ [1ex] 
 \hline
\end{tabular}
\vspace{1mm}
\caption{AUC (best method for each dataset bolded).}
\label{tab_exp_AUC}
\end{table}

\begin{table}[h!]
\centering
\begin{tabular}{|c c c c c|} 
 \hline
 & DBLP-3 & DBLP-5 & Reddit & Brain  \\ [0.5ex] 
 \hline\hline
 GCN-SE & \textbf{0.5766} & \textbf{0.6253} & 0.5056 & \textbf{0.9068} \\ 
 GCN-SE-1 & 0.5180 & 0.5725 & 0.5073 & 0.8954 \\ 
 RNNGCN & 0.5382 & 0.5540 & 0.5071 & 0.8569 \\
 GraphSage & 0.5365& 0.5757 & 0.5011 & 0.8987 \\
 GCN & 0.5543 & 0.5950 & 0.5213 & 0.8701 \\
 GC-LSTM & 0.5667  & 0.5393 & \textbf{0.5542} & 0.8908 \\ [1ex] 
 \hline
\end{tabular}
\vspace{1mm}
\caption{F1 score (best method for each dataset bolded).}
\label{tab_exp_F1}
\end{table}


\section{Attention as Explanation}\label{sec:explanation}


We next verify that our use of attention does in fact correlate with the importance of different graph snapshots in GCN-SE.
While attention weights intuitively should correlate with feature importance, some have argued that attention is not always a rigorous metric for explainability~\cite{serrano2019attention}. Thus, it is important to verify that the use of attention in GCN-SE does in fact correlate with the importance of different graph snapshots. 
%
We also demonstrate the utility of the proposed GCN-SE, by showing that it can explain the behavior of dynamic graph models in several practical application scenarios where the importance of graph snapshots is expected to vary over time.

\subsection{Verifying Attention-Importance Relationships}\label{subsec:experiment-verify}
In this section, we construct several experiment settings in which we know which timesteps are more important to the classification results, and show that the attention weights from GCN-SE match this prior knowledge. These investigations motivate our more formal evaluation of the correlation between attention and snapshot importance in the next section. 

To construct these experiments, we note that for dynamic graphs, the importance of a snapshot can be correlated to either the information from graph topology or node attributes. Thus, we can verify the relationship between attention and the importance of different snapshots by manually enforcing snapshots of the dynamic graph to have different relevance to the label to be predicted. To do so, we construct simulated data sets by first generating random labels for each node for the first time step. At each subsequent timestep, a node's label has 5\% chance to change to a different, randomly selected class. 
In each time step, edges form between nodes of the same label with probability 10\%, and between nodes of different labels with probability 0.5\%. 

\begin{figure}[ht]%
    \centering
    \includegraphics[width=6.5cm]{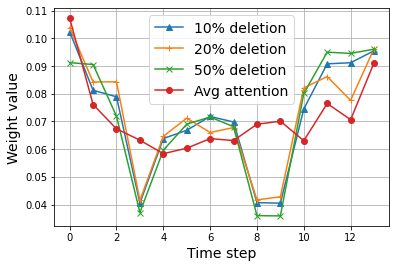}%
    \caption{Deleting 10\%, 20\% and 50\% edges on timestep 3, 8 and 9, resulting in decrease on attention weights of these time steps to different extents}%
    \label{fig_deleting_edges}%
\end{figure}

We would expect that \textbf{deleting some edges} from one or more snapshots would reduce the importance of those snapshots. Graphs without dynamic attributes in particular (e.g., our DBLP-E dataset, or the simulated dataset above) must predict node labels from the snapshot topologies, and changing the topology at a given time by deleting edges of that snapshot will disturb original information given by original data, which reduce its relevance to the prediction target. In Figure~\ref{fig_deleting_edges}, we apply edge deletion for time steps of 3, 8 and 9. Deletion results in apparent weight drops for the deleted time steps comparing to baseline weights, and the weights decrease more as we delete more edges.

%
%
%

\begin{figure}[ht]%
    \centering
    {\includegraphics[width=6.5cm]{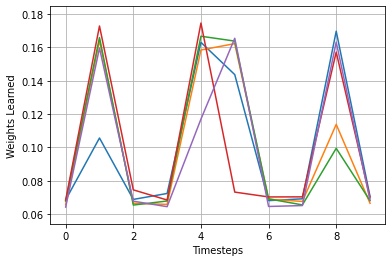} }%
    {\includegraphics[width=6.5cm]{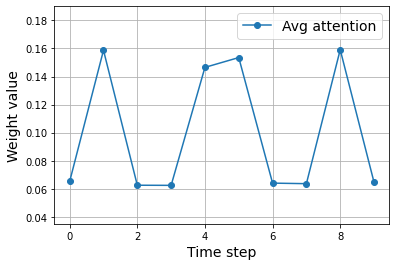} }%
    \caption{Specific time steps (1, 4, 5, 8) are assigned greater relevance to the label prediction. The GCN-SE attention weights correspondingly show consistently higher values on these time steps compared to others, over multiple runs (top) and averaged over these runs (bottom).}%
    \label{fig_veri_fixed_labels}%
\end{figure}

We now consider two other ways to enforce graph snapshots to have different importance. 
First, for datasets where node labels stay consistent for all time steps, we \textbf{assign higher connection probabilities} for nodes within specific classes in certain time steps, making them more relevant to the label prediction. In Figure~\ref{fig_veri_fixed_labels}, at time steps 1, 4, 5, and 8 we choose four classes whose nodes have a 40\% probability of forming intra-class edges and 10\% for inter-class edges. Thus, these classes are more densely connected and can be more readily classified based on information in these snapshots. As shown in Figure~\ref{fig_veri_fixed_labels}, the attention weights at time steps 1, 4, 5 and 8 are significantly larger than those for other time steps over multiple runs (top figure), as is their average (bottom figure), which aligns with our assigned importance. 

\begin{figure}[ht]%
    \centering
    {\includegraphics[width=6.5cm]{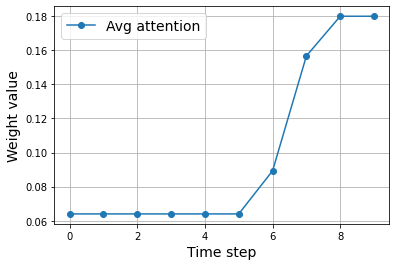} }%
    \caption{When node labels change over time, earlier time steps are less relevant to the prediction, which matches the attention weights learned by GCN-SE.}%
    \label{fig_veri_changing_labels}%
\end{figure}

Second, when node labels change over time, we would expect that \textbf{earlier snapshots} are less relevant to the final class predictions because of the randomness of label transition. Indeed, RNNGCN~\cite{yao2020interpretable} makes this assumption when using exponentially decaying weights across timesteps. 
We vary the labels over time in our simulated dataset and show the learned attention weights in Figure~\ref{fig_veri_changing_labels}: they increase for later timesteps, matching our intuition and approximating the exponential decay used by RNNGCN. 


\begin{figure}[ht]%
    \centering
    {\includegraphics[width=6.5cm]{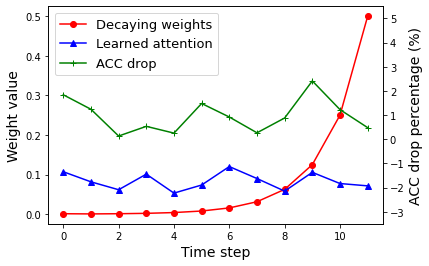} }%
    \caption{Decaying weights ($\lambda = 0.5$) from \protect\cite{yao2020interpretable}
    and learned attention weights from GCN-SE on the DBLP5 dataset.}%
    \label{fig_diff_weight}%
\end{figure}

\subsection{Measuring Attention-Importance Correlations}

The verification experiments in Figures~\ref{fig_deleting_edges} to~\ref{fig_veri_changing_labels} provide promising indications that GCN-SE's attention weights give some indication of snapshot importance. However, it is difficult to generalize them to real datasets, for which the importance of graph topology and node attributes can vary due to the model and data, and it may not be easy to manually enforce different snapshots to carry more or less information about the predicted label. Prior works on interpretable machine learning models suggest applying perturbations to or masking each input feature and measuring the resulting change in model accuracy~\cite{bhatt2020explainable} to evaluate its importance. We adapt these ideas by defining ``perturbations'' for our graph snapshots and examining the correlation between accuracy changes and GCN-SE's attention weights. For simplicity, we show results only for attention weights on dynamic topologies, though similar methods may be applied to weights for dynamic attributes.

Consider a dynamic graph with adjacency matrices $\mathbf{A} = \left( \mathbf{A}_1, \mathbf{A}_2, ..., \mathbf{A}_T \right)$, where $T$ is the total number of available time steps. We train a GCN-SE model and acquire the learned attention weights $\mathbf{W}$ through averaging multiple runs. To evaluate whether these weights are correlated with the snapshot importance, we use $\mathbf{W} = \left( W_0, W_1, ..., W_T \right)$ to take a linear combination of the dynamic feature maps of another baseline model, such as GCN. With the attention weights frozen, the baseline model is trained and we define the resulting accuracy as $m$. By setting the weight value at time step $k$ to be 0, $W_k = 0$, we get a new \emph{masked} attention $\mathbf{\Tilde{W}}_k = \left(W_1, ..., W_{k-1}, 0, W_{k+1}, ..., W_T \right)$. For all $0 \leq k \leq T$ and all new weights $\mathbf{\Tilde{W}}_k$ generated by masking each timestep, we separately retrain our baseline model while applying and freezing these masked weights in an attention layer. We denote the resulting accuracy as $m_k$, and the importance $I_k$ for timestep $k$ is defined by the decrease in accuracy:

\begin{equation}
    I_k = m - m_k.
\end{equation}
We repeat the masking and attention calculation operations above for at least 20 runs, so that the noises introduced by randomness in the training are neutralized. With importance defined as above, we can calculate the correlation coefficients between the learned attention weights $\mathbf{W}$ and importance vector $\mathbf{I} = \left( I_0, I_1, ..., I_T\right)$.  
Table~\ref{tab_eval_GCN} shows the resulting correlations from the DBLP-3, DBLP-5, Reddit, and Brain datasets. The snapshot importance and the attention weights have a positive correlation on all datasets, indicating that the attention weights are indicative of the importance of each graph snapshot to the prediction accuracy. As shown in Figure~\ref{fig_diff_weight}, which uses DBLP-5 as an example, the attention weights and defined importance are highly positively correlated.

For comparison, we also calculate the correlation between the importance and attention weights under RNNGCN's assumption of exponentially decaying weights for timesteps further in the past. We train the RNNGCN model and calculate each timestep's weights, then follow the same masking and evaluation procedure as above. Figure~\ref{fig_diff_weight} shows the different weights from GCN-SE and RNNGCN; as we would expect from Table~\ref{tab_eval_GCN}, the accuracy drop is visibly correlated with the learned attention.  Table~\ref{tab_eval_decaying} shows that under the exponential decay assumption, there is no obvious correlation between the explained importance and the perturbation impact. Indeed, the correlations are slightly negative across all datasets, indicating that these weights are not indicative of the different snapshots' importance.

\begin{table}[h!]
\centering
\begin{tabular}{|c c c c c|} 
 \hline
 & DBLP3 & DBLP5 & Reddit & Brain \\ [0.5ex] 
 \hline
 Correlation Coefficient & 0.7595 & 0.3907 & 0.3828 & 0.4808\\ [1ex] 
 \hline
\end{tabular}
\vspace{1mm}
\caption{Correlation coefficients between snapshot importance and average GCN-SE attention weights}
\label{tab_eval_GCN}
\end{table}

\begin{table}[h!]
\centering
\begin{tabular}{|c c c c c|} 
 \hline
 & DBLP3 & DBLP5 & Reddit & Brain \\ [0.5ex] 
 \hline
 Correlation Coefficient &  0.0341 & -0.2580 & -0.1041 & -0.0263\\ [1ex] 
 \hline
\end{tabular}
\vspace{1mm}
\caption{Correlation coefficients between snapshot importance and RNNGCN (decaying) weights}
\label{tab_eval_decaying}
\end{table}






\subsection{Applications of Attention as Explanations}

Having established that the GCN-SE attention weights are correlated with the importance of each graph snapshot, we finally introduce some scenarios where our proposed model could be used as a tool to explain the prediction decisions. Throughout this section, we consider the attention weights from a GCN-SE model trained to predict the node labels of the last time step of a dynamic graph. 

\textbf{Anomaly Detection.} 
\begin{figure}[t]%
    \centering
    {\includegraphics[width=6.5cm]{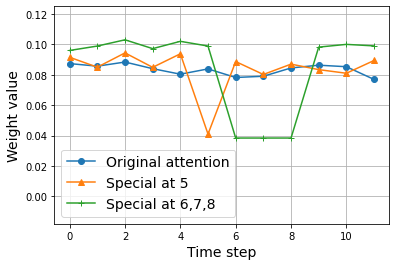} }%
    \caption{Attention results on manipulated data sets, case 1: nodes of time step 5 are assigned random labels; case 2: nodes of time step 6, 7, 8 are assigned random labels. The attention weights for these snapshots are lower than the baseline model for both cases, capturing their irrelevance.}%
    \label{fig_app_glitch}%
\end{figure}
\begin{figure}[t]%
    \centering
    {\includegraphics[width=6.5cm]{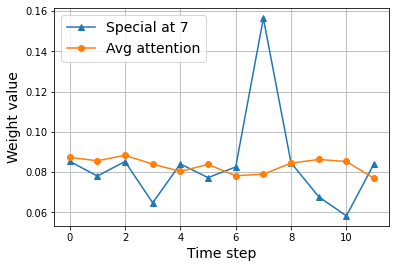} }%
    \caption{The most relevant or important time step is time step 7, and the learned attention weights correspondingly increase.}%
    \label{fig_app_oneistrue}%
\end{figure}
First, we can use the learned attention weights to locate the most relevant or the most irrelevant snapshots to the prediction target. For example, across different time steps there may be some irrelevant snapshots that should be treated as outliers, e.g., in the authorship graph, 2021 might be skewed by unusual collaboration patterns introduced by the COVID-19 pandemic. We test this application by considering the simulated dataset from Section~\ref{subsec:experiment-verify} and assigning random labels (and generating edges based on them) in a few selected timesteps. 
The snapshots in these timesteps are then irrelevant to the model prediction. Figure~\ref{fig_app_glitch} compares the resulting attention weights to the weights for the original graph, showing that they drop below average for these selected time steps. Conversely, we can also find the snapshot that is the most relevant to the label prediction. In Figure~\ref{fig_app_oneistrue}, only one snapshot is relevant to the prediction target, while other snapshots are assigned random labels to make them unimportant. The attention weight for this timestep is noticeably larger than the GCN-SE weights for the original graph, while the weights for other timesteps are similar to those for the original graph. 

\begin{figure}[t]%
    \centering
    {\includegraphics[width=6.5cm]{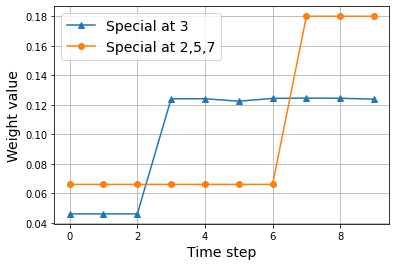} }%
    \caption{By setting higher label transferring probabilities on certain time steps, we expect attention weights can locate the last time step where this difference happens. For case 1, time step 3 is set higher and for case 2, time step 3, 5, 7 are set higher, we can see attention weights can successfully locate the pattern in this scenario}%
    \label{fig_stages}%
\end{figure}
\textbf{Detection of State Transitions.}
Suppose that at certain time steps, the topology of the snapshot and node labels change significantly and transition into another stable state. For example, the rapid rise of machine learning research might lead to a shift in collaboration patterns, which could produce a change of state in coauthorship graphs like DBLP-3 and DBLP-5. We can make use of attention to find the time steps where the snapshots begin to change. To test this idea, we change the node labels for different timesteps in our simulated dataset from Section~\ref{subsec:experiment-verify}, varying the probability that a node will change its label for different timesteps. Similar to the settings in Section~\ref{subsec:experiment-verify}, connections among nodes are generated based on labels.

Figure~\ref{fig_stages} shows the attention weights when time step 3 is set to have 80\% probability to change node labels while other time steps have a 20\% probability to change node labels. In another case, time steps 3, 5, and 7 have a high probability of changing node labels. We expect the attention weights to detect the last time step in which the probability of label changes is higher: at those timesteps, many nodes will change their labels, while fewer nodes will change their labels in other timesteps. Thus, only timesteps after the last set of changes will have edges generated based on labels that are largely similar to the labels in the last timestep. Figure~\ref{fig_stages}'s step-like attention weights reflect this expectation: there is a noticeable increase in the weights at time step 3 and time step 7 in our two cases.


\textbf{Periodicity Check.}
\begin{figure}[t]%
    \centering
    {\includegraphics[width=7.1cm]{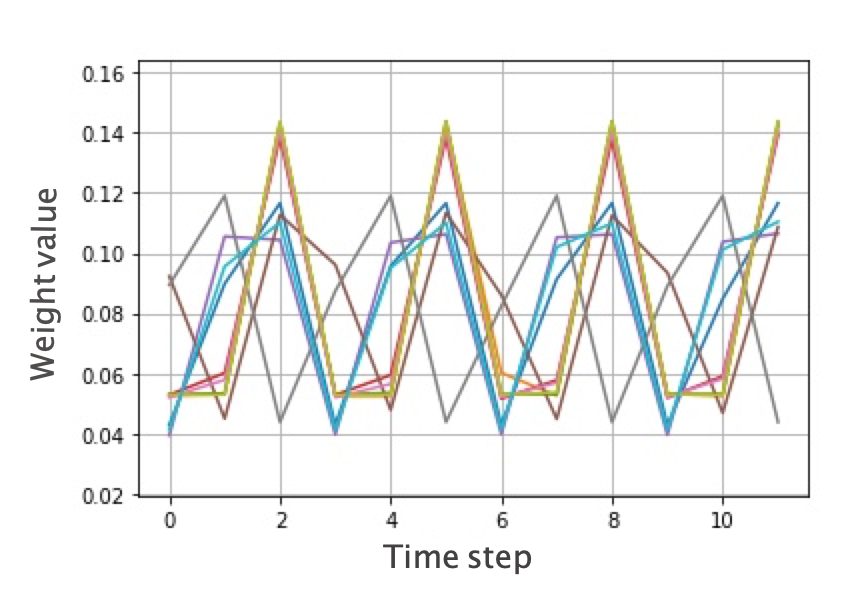} }%
    {\includegraphics[width=6.5cm]{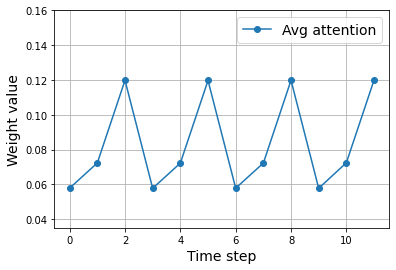} }%
   	\caption{Periodic attention weights learned on periodic dynamic graph data, created by modifying the Brain dataset. The top figure shows the attention weights from different runs, and the bottom figure shows the average weights. Both exhibit periodicity 3, matching the modified dataset.}%
    \label{fig_periodic}%
\end{figure}
%
%
Attention weights learned from the GCN-SE model can also be applied to datasets that exhibit temporal periodicity. For example, datasets of political activity on Twitter might exhibit yearly periodicity corresponding to U.S. presidential elections every four years. In such cases, the snapshots might show periodic patterns in their topology or node attributes over time. Unlike RNNGCN or the uniform aggregation of GCN, GCN-SE can flexibly learn attention weights that capture this periodicity. We can then examine the periodicity of the attention weights to detect the periodicity of the dataset itself. Figure~\ref{fig_periodic} shows the attention weights learned from GCN-SE on modified version of the Brain dataset, where snapshots are repeated every three timesteps with slight noise perturbations.  As shown in the top figure, each run shows a periodic pattern, which proves the robustness of GCN-SE. Although the attention weights for different runs exhibit different local minima, due to different sampling of the training and testing datasets, the bottom picture shows that a clear periodic pattern with periodicity of three timesteps.
